\documentclass[runningheads]{llncs}
\pdfoutput=1
\usepackage[T1]{fontenc}
\usepackage{graphicx}
\usepackage{acronym}

\usepackage{multirow}
\usepackage{booktabs}

\usepackage{pgfplots}
\usetikzlibrary{patterns}

\definecolor{mercury}{RGB}{240,240,240}
\definecolor{gallery}{RGB}{250,250,250}
\definecolor{free_speech_aquamarine}{RGB}{0, 156, 114}
\definecolor{shakespeare}{RGB}{35, 184, 223}
\definecolor{flamingo}{RGB}{237, 88, 85}
\definecolor{c1}{RGB}{255,97,56}
\definecolor{c2}{RGB}{67,114,196}
\definecolor{c3}{RGB}{237,126,50}
\definecolor{c4}{RGB}{165,165,165}
\definecolor{c5}{RGB}{254,192,4}

\AtBeginDocument{%
  \providecommand\BibTeX{{%
    \normalfont B\kern-0.5em{\scshape i\kern-0.25em b}\kern-0.8em\TeX}}}

\acrodef{LLM}{large language model}
\acrodef{RAG}{retrieval-augmented generation}
\acrodef{NLTK}{Natural Language Toolkit}
\acrodef{Open Domain QA}{Open Domain Question Answering}
\acrodef{LFQA}{Long-form Question Answering}
\acrodef{QUITO}{Query-gUIded aTtention cOmpression}
\acrodef{NLP}{natural language processing}
\acrodef{ICL}{in-context learning}
\acrodef{CoT}{chain-of-thought}
\acrodef{GQA}{Grouped-Query Attention}
\acrodef{SWA}{Sliding Window Attention}

\begin{document}
\title{QUITO: Accelerating Long-Context Reasoning through Query-Guided Context Compression}
\titlerunning{QUITO}
%
\author{Wenshan Wang\inst{1} \and
Yihang Wang\inst{2} \and
Yixing Fan\thanks{Corresponding Author: fanyixing@ict.ac.cn}\inst{1} \and
Huaming Liao\inst{1} \and
Jiafeng Guo\inst{1}}

\authorrunning{W. Wang et al.}
%
\institute{1. Institute of Computing Technology, Chinese Academy of Sciences \\
2. Beijing University of Posts and Telecommunications \\}
%
\maketitle              
\begin{abstract}

In-context learning (ICL) capabilities are foundational to the success of \acp{LLM}. 
Recently, context compression has attracted growing interest since it can largely reduce reasoning complexities and computation costs of \acp{LLM}.
In this paper, we introduce a novel \ac{QUITO} method, which leverages attention of the question over the contexts to filter useless information. Specifically, we take a trigger token to calculate the attention distribution of the context in response to the question. Based on the distribution, we propose three different filtering methods to satisfy the budget constraints of the context length.
We evaluate the \ac{QUITO} using two widely-used datasets, namely, NaturalQuestions and ASQA. Experimental results demonstrate that \ac{QUITO} significantly outperforms established baselines across various datasets and downstream \acp{LLM}, underscoring its effectiveness. Our code is available at \url{https://github.com/Wenshansilvia/attention\_compressor}.

\end{abstract}
\section{Introduction}
In recent years, \acp{LLM} has demonstrated notable reasoning and generating capabilities, significantly enhancing the performance of \ac{NLP} tasks~\cite{Brown2020}. 
However, these models still exhibit limitations in acquiring real-time information and integrating external knowledge~\cite{RAGSurvey}. 
\Ac{ICL} addresses these deficiencies by including examples and relevant contexts directly within the prompts\cite{dong2024surveyincontextlearning}. This approach boost the performance of \acp{LLM} in downstream tasks without requiring additional training.

To better improve the reasoning ability of \acp{LLM}, researchers propose different ways to incorporate complex contexts in the input~\cite{Brown2020,RAGSurvey}. For example, \ac{RAG} employs an additional searcher to retrieve external relevant documents about the question as the context of inputs, which has attracted lots of attention for both the academia and industry~\cite{asai2023selfraglearningretrievegenerate,borgeaud2022improvinglanguagemodelsretrieving,RAGSurvey}. In addition,
Brown et al.~\cite{Brown2020} found that the number of examples has a great impact to the reasoning performance of \acp{LLM}, where more examples tend to bring better performances \cite{song2022comprehensivesurveyfewshotlearning}. 
Moreover, the \ac{CoT} \cite{wei2023chainofthoughtpromptingelicitsreasoning,trivedi2023interleavingretrievalchainofthoughtreasoning} further improves the \acp{LLM} by involving the reasoning step of each example in the context. 
While these strategies have the potential to significantly improve the capabilities of \acp{LLM}, they also introduce challenges associated with the increased context length, such as higher inference complexity and costs.


To mitigate this issue, context compression in \ac{ICL} is becoming a prominent solution. On one hand, reducing the length by removing noise from contexts can improve inference efficiency\cite{SelectiveContext,xu2023compresspromptimprovingaccuracyefficiency}. On the other hand, it meets the input length restrictions of open-source LLMs\cite{touvron2023llamaopenefficientfoundation,yang2023baichuan2openlargescale} while also reduces the costs associated with accessing proprietary LLMs. 
Several methods~\cite{SelectiveContext,LLMLingua} have been proposed to compress context by estimating the information entropy. This assessment is conducted by utilizing a small external LLM to evaluate the perplexity of individual tokens to identify those that contribute minimal information gain. Tokens that demonstrate low information are subsequently compressed or eliminated. 
However, neglecting the query during compression may result in the inadvertent deletion of key information. 

For the above problem, recent methods such as LongLLMLingua \cite{LongLLMLingua} adopt a query-aware compression approach by calculating the perplexity of the context conditioned on the query. Despite this advancement, misalignment between compression model and generation model can lead to inconsistencies in determining which tokens are considered to have ``low entropy gain''. This discrepancy arises because models may differ in their interpretation and processing of the same information. Our work also scores tokens based on their relevance to the query. However, distinctively, we employ attention metrics rather than perplexity to assess the importance of tokens.

This paper introduces the \acf{QUITO} method, which strategically selects the context to maintain supporting information by utilizing the attention mechanism. 
Intuitively, the attention mechanism offers a direct method for analyzing the interactions between the question and the context, moving beyond the sole reliance on models' probabilistic uncertainty. This technique facilitates a more precise identification of the information that is most crucial to the current task. 
More importantly, the attention-based filtering can be implemented with small \acp{LLM}, which improves the computation efficiency.

\par The main contributions of this study include: 
\begin{enumerate}
    \item This paper proposes a novel context compression method, named \ac{QUITO}. It utilises self-attention mechanism of Transformers to score the importance of tokens, selecting context relevant to the current query. 
    \item In contrast to earlier methods that requires a compression model with 7 billion or 13 billion parameters, this method achieves superior results using a smaller \ac{LLM} with only 0.5 billion parameters.
    \item We conduct extensive experiments on two benchmark datasets, which demonstrate the effectiveness of the proposed \ac{QUITO}. For example, it surpasses strong baselines with an increase in accuracy of up to $20.2$.
\end{enumerate}

\section{Related Work}
In this section, we briefly review two lines of related works, i.e., context compression task and attention mechanism.

\subsection{Context Compression Task}
To reduce the length of context, earlier efforts\cite{RETA-LLM} opted to summarize and condense retrieved documents using models such as GPT. Other studies~\cite{LeanContext,RECOMP,FILCO,Fit-RAG} focused on distinguishing between useful and redundant information within documents, training a model to extract the most valuable sentences. For example, LeanContext~\cite{LeanContext} and FILCO~\cite{FILCO} train the model to perform sentence-level extraction for the context. Fit-RAG~\cite{Fit-RAG} scores sub-paragraphs with sliding context windows. RECOMP~\cite{RECOMP} uses a generative model to rewrite extracted candidate sentences, thereby ensuring the coherence and naturalness of the summaries.

Approaches that generate summaries do not allow direct control over the compression ratio, resulting in a growing attention on token and word-level compression techniques in recent times. SelectiveContext~\cite{SelectiveContext} utilizes self-information within context for token selection. This approach considers perplexity (PPL) to be the representation of the uncertainty of an LLM regarding information carried by contexts. 
Based on~\cite{SelectiveContext}, LLMLingua~\cite{LLMLingua} introduces a two-stage, coarse to a fine, compression method. 
However, these methods fail to consider the relationship between the context and the query. LLMLingua~\cite{LongLLMLingua} further addresses this gap by calculating context-specific perplexity conditioned on the query.

The aforementioned token-level compression methods utilize perplexity as the primary filtering criterion. However, discrepancies often arise between smaller compression models and larger generation models in their assessments of word perplexity, making it challenging to align their judgments on lexical importance.

\subsection{Attention Mechanism}
Attention is a significant breakthrough in deep learning, particularly shines in \ac{NLP} tasks such as translation and summary generation\cite{chaudhari2021attentivesurveyattentionmodels}. The core concept behind Attention mechanisms involves assigning a specific weight to each input element, such as words or tokens, indicating their relevance to the task at hand. This allows models to focus selectively on more pertinent parts of the input data.

Self-attention, a particular category of the attention mechanism, measures the relationships between all input elements, assessing how each element influences and relates to the others\cite{correia2021attentionpleasesurveyneural}. 
Multi-head attention is a key component of the Transformers~\cite{Vaswani2017}, which improves the model’s capability in capturing diverse correlation patterns.
Recent studies try to use the attention mechanisms within LLMs to accomplish specific tasks. For instance, DRAGIN~\cite{DRAGIN} use attention to evaluate the extent to which a given text segment significantly influences subsequent content.
It employs the perplexity of tokens to determine whether to trigger re-retrieval and regeneration processes.
In this paper, we also employ the multi-head attention mechanism to calculate the weights of tokens in context, thereby identifying useless content for answer generation.

\section{Method}

\begin{figure}[t!]
\includegraphics[width=1.0\textwidth]{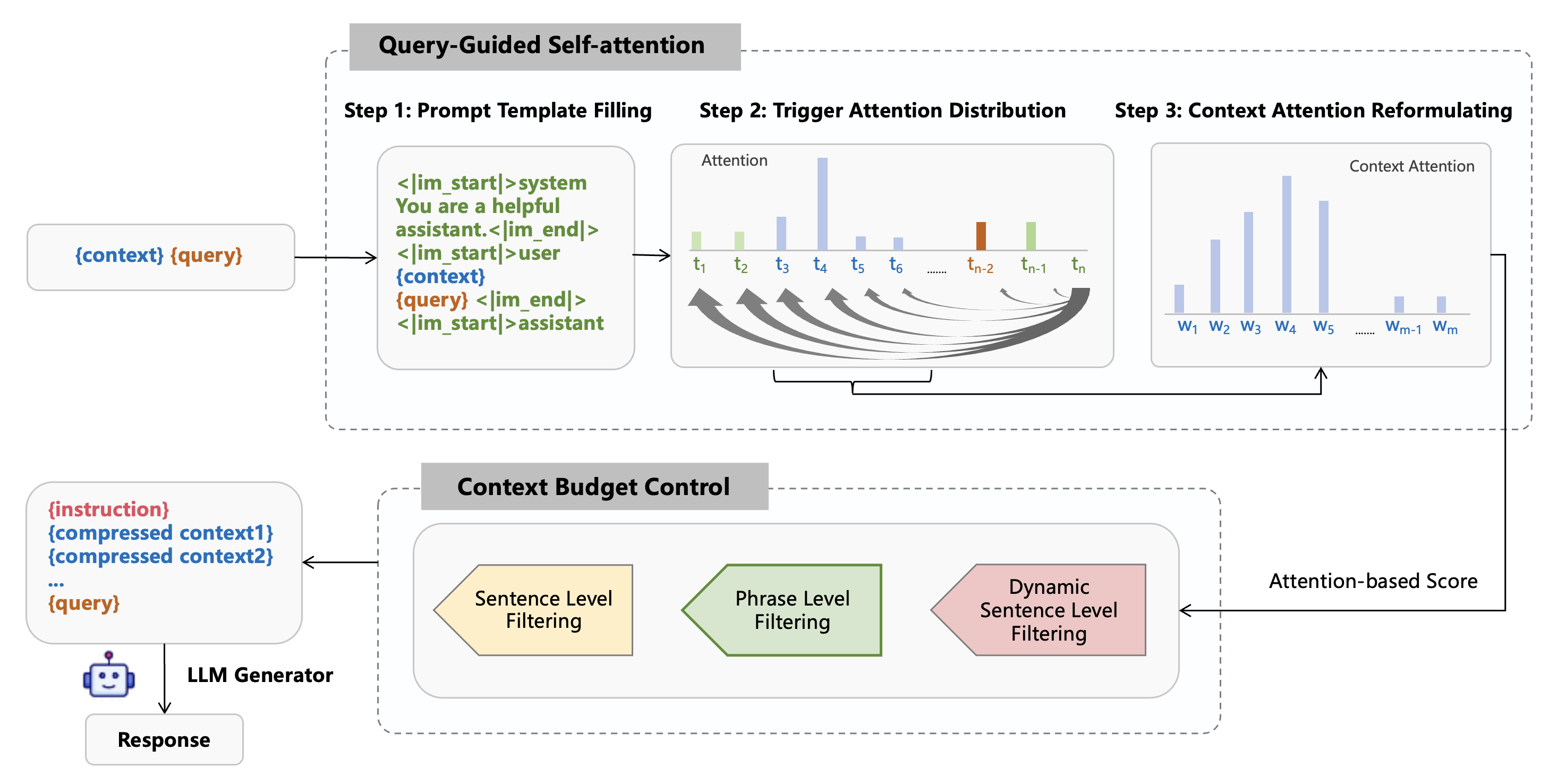}
\caption{The overall framework of \ac{QUITO}} 
\label{fig1}
\end{figure}

In this section, we introduce the \ac{QUITO} method in detail. As illustrated in Figure~\ref{fig1}, \ac{QUITO} primarily consists of two main components, namely \textit{the query-guided self-attention} component and \textit{the context budget control} component. 
In what follows, we will firstly give a formal definition of the task, and then describe each component in detail.

\subsection{Problem Formulation}
Given an input $p=(s, C, q)$, where $s$ is the instruction, $q$ is the query, and $C=\{c_i\}_{i=1}^{n}$ is the context consisting $n$ documents. Every document $c_i=\{w_{i, j}\}_{j=1}^{L_{i}}$ contains $L_{i}$ word. The objective of context compression task can be formulated as:
\begin{equation}
\min_{\tilde{C}}{dist(P(\tilde{y}|s, \tilde{C}, q), P(y|s, C, q))}, 
\end{equation}
where $\tilde{y}$ represents the predicted response of the LLM, and $y$ is the ground truth response. $dist ( \cdot , \cdot )$ is a function that measures the distance of two distributions, such as KL divergence. 
$\tilde{C}=\{\tilde{c_i}\}_{i=1}^{n}$ is the compressed context, and  $\tilde{c_i}=\{w_ j|w_j\in c_i\}_{j=1}^{\tau L_{i}}, \tau \in [0, 1]$. $\tilde{c_i}$ is $c_i$ being compressed with ratio at $1/\tau$, where smaller $\tau$ means higher compression ratio.

\subsection{Query-Guided Self-Attention}
The query-guided self-attention component aims to estimate the importance of tokens in context by calculating the trigger attention distribution. Firstly, we organize all input by \textit{prompt template filling}. Then, we calculate the importance of all the input with \textit{trigger attention distribution}. Finally, we obtain the lexical units importance within the context by \textit{context attention reformulating}.

\subsubsection{Prompt Template Filling}
It is crucial that the compression model fully understands the task at hand and accurately identifies the information most pertinent to the current query. A standard approach involves concatenating the context with the query and subsequently analyzing how tokens within the query attend to tokens in the context. However, in a Transformer decoder-only architecture, the visibility range of each token in the query varies. This variability suggests that tokens positioned later in the sequence more precisely reflect the model’s comprehensive understanding of the task. Given the challenges associated with appropriately weighting tokens at different positions, we propose a novel method that utilize a conversational template and identify a specialized token that encapsulates the compression model’s overall understanding of the task.

\subsubsection{Trigger attention distribution}
We embed the context and query into a conversational template, concluding with a signal that prompts the model to initiate response generation. The terminal token within this sequence is designated as a trigger token, serving as an indicator of the model’s assessment of information need after comprehensively understanding the task at hand. Subsequently, we employ a compression model equipped with a multi-head self-attention mechanism to process the completed template and compute the attention that the trigger token accords to the preceding text:
\begin{equation}
\{\alpha_i|\alpha_i=\frac{\exp(q_{L_{total}}^Tk_i)}{\sum_{j=1}^{L_{total}}{\exp(q_{L_{total}}^Tk_j})}\},
\end{equation}
where $q_i$ and $k_i$ are query embedding and key embedding of the $i th$ token, respectively. $L_{total}$ is the total number of tokens in the completed template. 

\subsubsection{Context attention reformulating}
Once the attention allocated by the trigger token to all preceding tokens in the sequence has been determined, the subsequent step involves transforming this attention data into a quantified measure of significance for the lexical units within the context. 
\par The array $\{\alpha_i\}$ signifies attention weights, with its length equating to the aggregate of the lengths of the conversational template, the context, and query. Within the scope of this task, it is imperative to concentrate on the attention distributed to the context segment. The attention should not be diluted by the segments pertaining to the template and the query. Consequently, we implement a normalization process, which is designed to ensure that the distribution of attention across various tokens in the context remains unbiased, robust to the disparities in context and query lengths that may exist across different tasks. For the normalization we use softmax function:
\begin{equation}
    \alpha_i'=\frac{\exp(\alpha_{i+doc_{start}})}{\sum_{j=doc_{start}}^{doc_{end}}{\exp(\alpha_j)}}, i \in[1, doc_{end}-doc_{start}], 
\end{equation}
where $doc_{start}$ and $doc_{end}$ represent the starting and ending positions, respectively, of the context segment.

We consider words to be the smallest semantic units within a document. In order to perform selection on semantic units, the next step involves transforming scores on token to scores attributed to each individual unit. In other words, we need to transform $\{\alpha_i'\}_{i=1}^{L_{doc}}$ to $\{\alpha_i''\}_{i=1}^{L}$, where $L_{doc}$ is the length of token array $\{t_i\}_{i=1}^{L_{doc}}$ that belongs to context, and $L$ is the length of word array $\{w_i\}_{i=1}^{L}$.
\par A word $w_i$ may consist of one or more tokens. We can formulate a word as $w_i=\{t_j\}_{j=k+1}^{k+l}$, each of which has attention score:
\begin{equation}
    \alpha_i''=\max_{k+1 \leq j \leq k+l}{\alpha_j'},
\end{equation}
where the length of the array $\{\alpha_i''\}_{i=1}^{L}$ is $L$.

\subsection{Context Budget Control}
In the previous section, we have derived a list of words, represented as $\{w_i\}_{i=1}^{L}$, and the corresponding array of attention weights, $\{\alpha_i''\}_{i=1}^{L}$. 
This section introduces the filtering methods that satisfy the requirement of the context budget control.

\subsubsection{Phrase Level Filtering}
In the process of selecting based on attention scores, it is common to inadvertently overlook words adjacent to those with high attention, referred to as target words, which may also contain crucial knowledge for answering the query. To rectify this oversight and ensure these adjacent words are also considered, we apply a weighted adjustment, allowing them to receive a portion of the attention attributed to the target words. This is accomplished by implementing a Gaussian filter across the word attention array $\{\alpha_i''\}_{i=1}^{L}$.
\begin{equation}
    G(x)=\frac{1}{2\pi\sigma^2}\exp(-\frac{x^2}{2\sigma^2})
\end{equation}
After the application of the Gaussian function $G(x)$ to $\{\alpha_i''\}_{i=1}^{L}$, the resulting Gaussian-modulated attention array is denoted as $\{\alpha_i'''\}_{i=1}^{L}$.
\par Subsequently, we identify the words from set $\{w_i\}_{i=1}^{L}$ that rank within the top $\tau L$ based on their attention scores $\{\alpha_i'''\}_{i=1}^{L}$.
\begin{enumerate}
    \item Perform a sort on $\{\alpha_i'''\}_{i=1}^L$ on descending order, which yields an ordered set of indices $\{j_1, j_2, \ldots, j_L\}$.
    \item Select corresponding words from $\{w_i\}_{i=1}^{L}$ with index $\{j_1, j_2, \ldots, j_{\tau L}\}$ ,which yields $\{w_i'\}_{i=1}^{\tau L}$.
    \item Reorganize the set of selected words $\{w_i'\}_{i=1}^{\tau L}$ to reflect their original sequential order within the context.
\end{enumerate}
\par Although the selection process targets individual words, the application of Gaussian filtering often leads to the selection of contiguous words, thereby effectively forming phrases.
\subsubsection{Sentence Level Filtering}
In addition to phrase-level filtering, sentence-level filtering is also implemented to preserve more comprehensive semantic information. Using the \ac{NLTK} toolkit, we extract semantic units at the sentence level. Each sentence $s_i$, denoted as 
$s_i=\{w_j\}_{j=k+1}^{k+l}$, is assigned an attention score based on the maximum score of the tokens it contains. Subsequently, mirroring the phrase-level filtering process, we prioritize incorporating sentences with higher attention scores into the selection set, while ensuring that the aggregate word count remains below $\tau L$.

\subsubsection{Dynamic Sentence Level Filtering}
Sentence-level filtering often leads to a compression ratio greater than the designated target $1/\tau$. To more effectively adhere to the predetermined compression rate and optimize budget utilization, we augment the results of sentence-level filtering with word-level filtering. Specifically, subsequent to sentence-level filtering, if the count of words is $L'$, we are then able to select an additional $\tau L-L'$ words. These additional words are chosen via phrase-level filtering from the text that was not previously selected. The newly selected words are subsequently concatenated with the results from sentence-level filtering to form the final compressed output.

\section{Experiments}

\subsection{Datasets and Evaluation Metrics}
In this paper, we assess the efficacy of the proposed \ac{QUITO} method across two distinct scenarios: open domain question answering and long-form question answering. Specifically, we employ the NaturalQuestions (NQ) and ASQA datasets as the testbed.

For NQ dataset, We employed a processed version as described in~\cite{lostinthemoddle}, where each query is paired with 20 documents, among which only one document contains the correct answer. In alignment with the procedures specified in~\cite{lostinthemoddle}, accuracy was used as the metric to determine whether the generated responses accurately included the correct answer.
For the ASQA dataset, the answer to the question maybe multi-facet as there are many ambiguous questions.
Each ambiguous question in the ASQA dataset has answers reflecting multiple interpretations of these ambiguities. We utilize the dataset version provided by~\cite{ALCE}, which includes 5 retrieved documents/snippets from Wikipedia for each query.
In accordance with~\cite{asqa}, our evaluation metrics included Exact Match (EM), a RoBERTa-based QA score (DisambigF1), and ROUGE~\cite{rouge}.


\begin{table}[tb]
    \centering
    \setlength{\tabcolsep}{1mm}
    \resizebox{\columnwidth}{!}{
    \begin{tabular}{l|c|ccc}
    \toprule
        \multirow{2}{*}{Methods} &  \multicolumn{1}{@{}c}{{\bf NQ}} &  \multicolumn{3}{@{}c}{{\bf ASQA}} \\
        & Accuracy & RougeL & EM & Disambig\_F1\\
    \midrule
    \midrule
    \multicolumn{5}{@{}c}{{ \textit{ratio=2x}}} \\
    \midrule
    Selective-Context & 53.2 & - & - & - \\
    LLMLingua & 38.7 & 21.3 & 34.6 & 22.2\\
    LongLLMLingua$\dagger$ & 41.2 & 21.6 & 29.7 & 21.2\\
    \midrule
    {QUITO (Sentence Level)} & 49.9 & 23.5 & \textbf{40.3} & 23.6 \\
    {QUITO (Dynamic Sentence Level)} & 58.3 & \textbf{23.5} & 40.0 & \textbf{23.8}\\
    {QUITO (Phrase Level)} & \textbf{58.9} & 21.6 & 38.3 & 22.8 \\
    \midrule
    \midrule
    \multicolumn{5}{@{}c}{{ \textit{ratio=4x}}} \\
    \midrule
    Selective-Context & 38.2 & - & - & -\\
    LLMLingua & 32.1 & 20.9 & 33.2 & 21.1\\
    LongLLMLingua$\dagger$ & 33.6 & 20.9 & 24.2 & 20.2 \\
    \midrule
    {QUITO (Sentence Level)} & 52.1 & 22.1 & 30.1 & 20.2 \\
    {QUITO (Dynamic Sentence Level)} & \textbf{53.1} & \textbf{22.5} & \textbf{36.7} & \textbf{22.5}\\
    {QUITO (Phrase Level)} & 50.7 & 20.8 & 34.7 & 21.5 \\
    \midrule
    \midrule
    Original (without compression) & 68.6 & 23.0 & 45.7 & 26.2 \\
    \bottomrule
    \end{tabular}
    }
    \caption{Experimental results of various compression methods applied at different compression ratios on the NaturalQuestions and ASQA datasets. 
    }
    \label{tab:main-result}
\end{table}

\subsection{Baselines and Implementation}
\subsubsection{Baselines}
We take three state-of-the-art compression approaches as baselines: For query-unaware methods, we select Selective-Context\cite{SelectiveContext} and LLMLingua\cite{LLMLingua}, which implements cross entropy scoring to remove redundant vocabulary. For query-aware method, we compare our approach with Longllmlingua\cite{LongLLMLingua}. LongLLMLingua implements a two-stage compression method. It first evaluates and reranks multiple retrieved contexts, followed by a token-level compression stage, allocating varying compression budgets to these contexts based on their initial scores. 
For fair comparison, we excluded the context reranking phrase of LongLLMLingua (marked as LongLLMLingua$\dagger$ in Table \ref{tab:main-result} and Figure~\ref{contextposition}), concentrating on the token-level compression.


\subsubsection{Detailed Implementation}
For fair comparison, we follow LLMLingua~\cite{LLMLingua} to use Longchat-13B-16k~\footnote{https://huggingface.co/lmsys/longchat-13b-16k} as the generation model.
To ensure the reproducibility of the results, we apply greedy decoding strategy throughout the inference process, with the temperature parameter set to zero. The compression model is implemented with Qwen2-0.5B-Instruct\footnote{https://huggingface.co/Qwen/Qwen2-0.5B-Instruct}.

\subsection{Main Results}

Table \ref{tab:main-result} presents the comparative performance of our method, \ac{QUITO}, against three baseline methods across various compression rates and datasets. 
Firstly, we can see that selective-context is a strong baseline compared with LLMLingua and LongLLMLingua$\dagger$ on both $2x$ and $4x$ compression rates.
Secondly, \ac{QUITO} obtains significantly better performances than all baselines, e.g., the improvement of \ac{QUITO} with phrase level filtering against selective-context, LLMLingua, and LongLLMLingua$\dagger$(i.e., $2x$ compression ratio) on NQ is $5.7$, $20.2$, and $17.7$, respectively. 
Finally, we find that \ac{QUITO} with different filtering method all achieve better performances on both datasets. However, there is no consistent advantages of each filtering method when compared on different datasets. This maybe that the context length on NQ and ASQA differs significantly, i.e., the average length of context on NQ and ASQA is about $2904$ and $721$ tokens, respectively.
All the results demonstrate the effectiveness of \ac{QUITO} in compressing contexts for the \acp{LLM}.


\subsection{Analysis on different position of the ground truth context}
We analyse the performance of the \ac{QUITO} compression method across different ground truth context positions within the NQ dataset. This dataset comprises 20 context document fragments per query, of which only one contains the answer and is designated as the ground truth document. We assessed the impact of this document's positioning at the 1st, 5th, 10th, 15th, and 20th ranks on the efficacy of various compression strategies. 



\definecolor{color8}{RGB}{169,209,142}
\definecolor{color7}{RGB}{84,130,53}
\definecolor{color4}{RGB}{244,177,131}
\definecolor{color3}{RGB}{237,125,49}
\definecolor{color6}{RGB}{157,195,230}
\definecolor{color5}{RGB}{46,117,182}
\definecolor{color1}{RGB}{139, 137, 112}
\definecolor{color2}{RGB}{205, 201, 165}


\pgfplotsset{
axis background/.style={fill=white},
grid=both,
  xtick pos=left,
  ytick pos=left,
  tick style={
    major grid style={style=gallery,line width=1pt},
    minor grid style=mercury,
    },
  minor tick num=1,
}

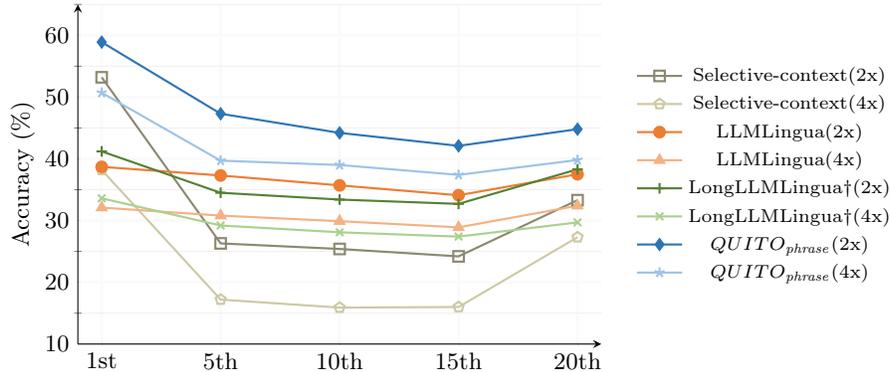
\begin{figure}[!tb]
\centering
  \begin{tikzpicture}
    \begin{axis}[
      height=.50\textwidth,
      width=.7\textwidth,
      xticklabels={1st, 5th, 10th, 15th, 20th},
      xtick={1,2,3,4,5},
      legend style={
          font=\scriptsize,
          draw=none,
          legend columns=1,
          at={(1.05,0.5)},
          anchor=west,
          /tikz/every even column/.append style={column sep=0.9mm}
        },
        ymajorgrids={true},
        ymin=10,
        ymax=65,
        ytick={10, 20, ..., 70},
        minor x tick num={0},
        minor y tick num={1},
        axis lines=left,
        enlarge x limits=0.05,
        tickwidth=0pt,
        legend entries = {Selective-context(2x), Selective-context(4x), LLMLingua(2x), LLMLingua(4x), LongLLMLingua$\dagger$(2x), LongLLMLingua$\dagger$(4x), $QUITO_{\mathit{phrase}}$(2x),$QUITO_{\mathit{phrase}}$(4x), Original prompt},
        ylabel={Accuracy (\%)},
        ylabel style={
      	anchor=south,
            at={(ticklabel* cs:0.5)},
            yshift=-22pt
        },
        ]
\addplot[color=color1,mark=square,
          thick
          ] coordinates {
        (1, 53.2)
        (2, 26.3)
        (3, 25.4)
        (4, 24.2)
        (5, 33.3)
      };
      \addplot[color=color2, mark=pentagon,
          thick
          ] coordinates {
        (1, 38.2)
        (2, 17.2)
        (3, 15.9)
        (4, 16)
        (5, 27.3)
      };
      \addplot[color=color3, mark=*,
      	 thick] coordinates {
        (1, 38.7)
        (2, 37.3)
        (3, 35.7)
        (4, 34.1)
        (5, 37.5)
      };
    \addplot[color=color4, mark=triangle*,
      	 thick] coordinates {
        (1, 32.1)
        (2, 30.8)
        (3, 29.9)
        (4, 28.9)
        (5, 32.4)
      };
      \addplot[color=color7, mark=+,
      	 thick] coordinates {
        (1, 41.2)
        (2, 34.5)
        (3, 33.4)
        (4, 32.7)
        (5, 38.3)
      };
    \addplot[color=color8, mark=x,
      	 thick] coordinates {
        (1, 33.6)
        (2, 29.2)
        (3, 28.1)
        (4, 27.4)
        (5, 29.7)
      };
    \addplot[color=color5, mark=diamond*,
      	 thick] coordinates {
        (1, 58.9)
        (2, 47.3)
        (3, 44.2)
        (4, 42.1)
        (5, 44.8)
      };
    \addplot[color=color6, mark=star,
      	 thick] coordinates {
        (1, 50.7)
        (2, 39.7)
        (3, 39)
        (4, 37.4)
        (5, 39.8)
      };
      
    \end{axis}
  \end{tikzpicture}
  \caption{Experimental comparison of different ground-truth context positions.}
  \label{contextposition}
\end{figure}

\par The results presented in Figure~\ref{contextposition} indicate that all context compression methods struggle with the 'lost in the middle' phenomenon, as described by \cite{lostinthemoddle}. Performance is optimal when the ground truth context is positioned at the beginning; however, it deteriorates significantly when the ground truth context is placed in the middle. Among the evaluated methods, LLMLingua\cite{LLMLingua} exhibits the most resilience to the 'lost in the middle' phenomenon. This robustness may be attributed to its strategy of allocating higher compression ratios to contexts containing a greater density of information. Overall, the \ac{QUITO} method consistently surpasses the two baseline methods across a variety of ground truth context positions and compression rates. On average, \ac{QUITO} improves upon the performance of Selective Context\cite{SelectiveContext} by +19.6 and LLMLingua\cite{LLMLingua} by +13.6.

\subsection{Analysis on different generation models}

\begin{figure}[tb]
\centering	
\begin{tikzpicture}
	\begin{axis}[
		ybar,
		axis on top,
        height=.42\textwidth,
        width=.7\textwidth,
        bar width=0.45cm,
        ytick={45,50,55,60,65,70,75,80},
        ymajorgrids, tick align=inside,
        major grid style={draw=white},
        minor y tick num={1},
        enlarge y limits={value=.1,upper},
        ymin=45, ymax=80,
        axis lines=left,
        enlarge x limits=0.25,
        legend style={
            at={(1.01,0.45)},
            font=\scriptsize,
            anchor=west,
            draw=none, 
            legend columns=1,
            /tikz/every even column/.append style={column sep=0.2cm}
        },
        legend cell align={left},
        ylabel={Accuracy},
        ylabel style={
            anchor=south,
            at={(ticklabel* cs:.56)},
            yshift=-20pt
        },
        symbolic x coords={
           Original prompt, 2x compression, 4x compression},
       xtick=data,
       nodes near coords={
        \pgfmathprintnumber[fixed zerofill,precision=1]{\pgfplotspointmeta}
       },
       every node near coord/.append style={anchor=south, font=\fontsize{6pt}{4pt}\selectfont},
	]
    \addplot [draw=none, fill=c2, postaction={pattern=north east lines, pattern color=white}] coordinates {
      (Original prompt,68.6)
      (2x compression, 58.9)
      (4x compression, 50.7)
      };
    \addplot [draw=none, fill=c5, postaction={pattern= crosshatch dots}, pattern color=white] coordinates {
      (Original prompt, 71.7)
      (2x compression, 62.3)
      (4x compression, 57.8)
      };
   \addplot [draw=none,fill=c3, postaction={pattern=vertical lines}, pattern color=white] coordinates {
      (Original prompt,75.3)
      (2x compression, 66.4)
      (4x compression, 58.5)
      };
   \addplot [draw=none, fill=c4, postaction={pattern=grid}, pattern color=white] coordinates {
      (Original prompt, 78.3)
      (2x compression, 64.5)
      (4x compression, 59.2)
      };
    
    \legend{LongChat-13B-16k,Llama3-8B-Instruct,GLM4-9b-chat,Mistral-7B-Instruct}
	\end{axis}

\end{tikzpicture}
\caption{Experimental results of different generation models on NQ dataset.}
\label{generationmodel}
\end{figure}
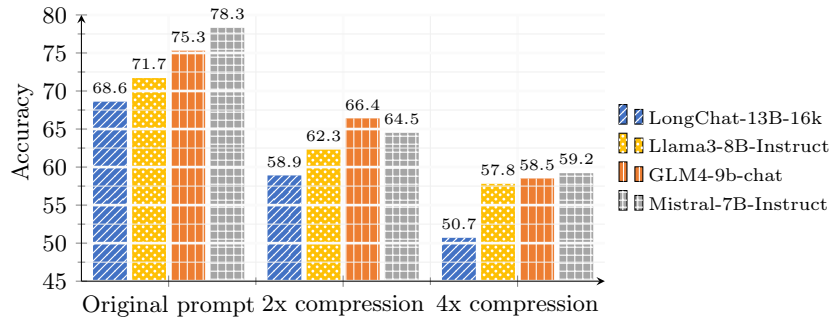

To better understand the generation ability of different \acp{LLM}, we evaluate the performance of 4 widely-used models, including Longchat-13B-16k~\footnote{https://huggingface.co/lmsys/longchat-13b-16k}, Llama3-8b-Instruct~\footnote{https://huggingface.co/meta-llama/Meta-Llama-3-8B-Instruct}, GLM4-9b-chat~\footnote{https://huggingface.co/THUDM/glm-4-9b-chat}, and Mistral-7b-instruct~\footnote{https://huggingface.co/mistralai/Mistral-7B-Instruct-v0.2}. These models were tested with contexts compressed at a rate of 2 on the NQ dataset. The generated responses from these compressed contexts were then compared with those derived from uncompressed contexts.

As depicted in Figure~\ref{generationmodel}, the Mistral-7B-Instruct model significantly outperforms the other three generation models despite having fewer parameters. This superior performance may be attributed to the incorporation of \ac{GQA} and \ac{SWA} during its training phase, which enhances its capability to process long sequence inputs.
While the context is compressed at 2x ratio, we find that the GLM4-9b-chat model show the smallest performance decline, with a decrease of $8.9$, and the Mistral-7B-Instruct has the greatest decline.
When the compression ratio is 4x, we can see that all generation models obtain a relative close performance except for LongChat-13B-16k. These maybe that the LongChat-13B-16k is released earlier than other three models, and the latter are trained more deeply. 

\section{Conclusion}
This paper introduces the \ac{QUITO} method, a novel attention-based importance estimation for long context compression in \acp{LLM}.
The \ac{QUITO} method employs a trigger token that comprehensively considers the query to assess the importance of each lexical unit within the context, thereby filtering out units with low relevance scores. 
Evaluations conducted on the NQ and ASQA datasets demonstrate that our method outperforms state-of-the-art compression methods such as Selective Context, LLMLingua, and LongLLMLingua, confirming its superior ability to preserve essential information needed by LLMs to respond to queries effectively.
For future work, we would like to study the combination of the context compression and re-ranking module, since the re-ranking stage in RAG also targets on selecting useful information for final answer generation.

%

%
%
%
\bibliographystyle{splncs04}
\bibliography{mybibliography}
%






\end{document}